\documentclass[12pt, sn-mathphys-num]{sn-jnl}

\usepackage{graphicx}%
\usepackage{multirow}%
\usepackage{amsmath,amssymb,amsfonts}%
\usepackage{amsthm}%
\usepackage{mathrsfs}%
\usepackage[title]{appendix}%
\usepackage{xcolor}%
\usepackage{textcomp}%
\usepackage{manyfoot}%
\usepackage{booktabs}%
\usepackage{algorithm}%
\usepackage{algorithmicx}%
\usepackage{algpseudocode}%
\usepackage{listings}%
\usepackage{soul}
\usepackage{makecell}
\usepackage{hyperref}
\usepackage{graphicx}
\usepackage{multicol}
\usepackage{color}
\usepackage{amsmath}
\usepackage{amssymb}
\usepackage{adjustbox}
\usepackage{caption}
\usepackage{titlesec}
\titlespacing*{\section}{0pt}{0.7\baselineskip}{0.7\baselineskip}
\titlespacing*{\subsection}{0pt}{0.5\baselineskip}{0.5\baselineskip}
\usepackage{geometry}
\usepackage{anyfontsize} 
\usepackage{setspace}    

\fontsize{12}{15}\selectfont
\begin{document}
\large
\title[Article Title]{Deep Time Warping for Multiple Time Series Alignment}


\author[1]{\fnm{Alireza} \sur{Nourbakhsh}}\email{alireza.nourbakhsh@ee.sharif.edu}

\author*[1]{\fnm{Hoda} \sur{Mohammadzade}}\email{hoda@sharif.edu}

\affil[1]{\orgdiv{Department of Electrical Engineering}, \orgname{Sharif University of Technology}, \orgaddress{\city{Tehran}, \country{Iran}}}


\abstract{Time Series Alignment is a crucial task in signal processing with wide-ranging applications. Real-world signals often suffer from temporal shifts and scaling, leading to errors in raw data classification. This paper presents a novel Deep Learning-based approach for Multiple Time Series Alignment (MTSA). Unlike existing methods, which mainly focus on Multiple Sequence Alignment (MSA) for biological sequences, there is a notable lack of alignment techniques for numerical time series. Traditional methods also typically address pairwise alignment, whereas our approach aligns all signals simultaneously, improving both alignment efficiency and computational speed. By decomposing signals into piece-wise linear sections, we introduce varying complexity into the warping function while ensuring compliance with three key constraints: boundary, monotonicity, and continuity conditions. Leveraging a deep convolutional network, we propose a new loss function that overcomes some limitations of Dynamic Time Warping (DTW). Experiments on the UCR Archive 2018, involving 129 time series datasets, show that our method significantly enhances classification accuracy, warping average, and runtime efficiency across most datasets.}




\maketitle

\section{Introduction}\label{sec1}

Multiple Sequence Alignment (MSA) and Multiple Time Series Alignment (MTSA) are essential in machine learning, data analysis, and bioinformatics, both aiming to align multiple inputs to identify patterns. The key difference lies in the data type: MSA aligns symbolic, discrete sequences like DNA, RNA, or proteins, while MTSA aligns continuous numerical signals, such as time series representing temporal or spatial measurements.

Both MSA and MTSA achieve alignment through a series of pairwise alignments. However, MTSA's numerical nature and higher computational complexity have restricted research in this area, whereas MSA has received extensive attention in the literature.

This paper addresses the research gap in MTSA by employing a multiple alignment algorithm instead of pairwise alignments, which leads to a better performance. Given the strong conceptual and methodological links between MSA and MTSA, we also review MSA approaches in the literature to gain insights for advancing MTSA methods.

The problem involves aligning a set of time series with arbitrary lengths. Due to its importance and wide applications, various approaches have been proposed for MSA and MTSA. At the heart of these methods is Dynamic Time Warping (DTW), the most widely used technique for signal alignment. 

In the following subsections, we present various applications of MSA and methods grounded in DTW.

\subsection{Applications}
The applications of MSA and MTSA can be categorized as follows:

\textbf{Classification:} Time series classification presents challenges due to shifts and rescaling in similar signals. A proper pre-warping stage can improve accuracy, as shown in the Experiments section. Studies \cite{tam,shape,stsad} have combined DTW and its extensions with Nearest Neighbor (NN) for classification, but DTW+NN requires computing DTW between each test and training sample. The Nearest Centroid (NC) algorithm \cite{scge} reduces this by aligning test samples with a representative signal per class, with \cite{ncc} further refining this into a classifier. Selecting the representative signal is crucial, commonly performed using Dynamic Barycenter Averaging (DBA) \cite{dba}, which iteratively aligns and updates the barycenter. Instead, we employ MTSA algorithms, achieving superior quality and performance, as demonstrated in the Experiments section.

\textbf{Human Activity Recognition:} HAR is a specialized classification task involving motion signals, widely used in surveillance, healthcare, assistive robotics, and human-machine interfaces. Here signal alignment is crucial due to variations in speed and initial phase across individuals performing activities like running or walking. Several time warping-based methods for HAR have been proposed in \cite{sjot,gtw,tam,ttn,ntw,fardl}.

\textbf{Biological Signal Analysis:} Signals such as ECG, EEG, EMG, and PPG serve as the primary channels in an intelligent system aimed at understanding human health situations. Due to variations in amplitude and morphology among biological signals, the absence of labeled datasets, and the difficulty of labeling, even by experts, the development of unsupervised warping approaches becomes imperative. Authors in \cite{beatlex} employ an algorithm based on DTW to identify sub-patterns in signals, utilized for signal prediction. In \cite{goa} and \cite{fecg}, DTW is applied to eliminate unwanted noise from ECG signals. Additionally, \cite{adtw} approximates DTW using a neural network on EEG signals.

Recently DTW and alignment methods have also been used for applications such as video alignment \cite{vid1, vid2, beatlex} and time series forecasting \cite{for1, for2}. While there are numerous other applications for MSA and MTSA in various domains, we omit them here for the sake of brevity.

\subsection{Methods}
MSA is widely used in genomics, particularly for protein sequence analysis, leading to the development of numerous methods in this field. The first method discussed is ClustalW \cite{clusteralw}. It performs pairwise alignment between signals to build a guide tree based on Progressive Alignment \cite{psa}, which assumes that aligning two similar signals allows them to be treated as one. Through iterative pairwise alignment, a set of time series can be aligned, but the signals need to be homogeneous, such as motion or ECG signals.

Hidden Markov Model (HMM) is used for MSA in literatures like \cite{deepmsa, hmmerge,witch}. In \cite{cpm}, an unsupervised approach models each time series as a non-uniformly distributed sample from a latent trace, accounting for local rescaling and noise. For MTSA, alignment is conducted separately using DTW between each signal and the latent trace. Notably, \cite{cpm} is one of the few works directly addressing numerical time series in MTSA.

In all the aforementioned works, Multiple Alignment is achieved through a series of pairwise alignments. Additionally, some studies like \cite{saga}, propose a method for aligning two signals and then extend it to MSA by aligning each signal with the average signal.

\section{Background} \label{sec2}
This section covers key concepts of MTSA, starting with warping and its definitions. It then explores DTW as the most common warping method, discusses its limitations, and presents novel approaches derived from it. Finally, the section outlines our contributions to the field.

\subsection{Overview of Useful Definitions} \label{sec2-1}
\textbf{Warping}: Consider two time series $X$ and $Y$ with lengths $N$ and $M$, respectively. The \emph{warping path}, denoted as $P$, is a sequence with length $L\in\mathbb{N}$ defined as follows:
\begin{equation}
\label{eq1}
P=(p_1,...,p_l)
\end{equation}
In Eq. \ref{eq1} for $l\in[1:L]$ we have $p_l=(n_l,m_l)\in[1:N]\times[1:M]$. Clearly $L=\max(N,M)$ and $p_l=(n_l,m_l)$ signifies that the index $n_l$ from $X$ is warped to index $m_l$ from $Y$. Thus, the warping path encapsulates all the necessary information for aligning the two signals. Typically, three \emph{warping constraints} are considered:
\begin{itemize} \itemsep0em \large
\item \emph{Boundary condition}: $p_1=(1,1)$ and $p_L=(N,M)$. This ensures that the first and last indices from the signals are warped to each other. 
\item \emph{Monotonicity condition}: $n_1\leq n_2\leq...\leq n_L$ and $m_1\leq m_2\leq...\leq m_L$. The alignment must preserve the chronological order of the time series.
\item \emph{Continuity condition}: $p_{l+1}-p_l\in\{(1,0),(0,1),(1,1)\}$ for each $l\in[1:L]$. This condition eliminates any jumps in finding corresponding points in the two signals, ensuring that all time steps have at least one corresponding point from the other signal. 
\end{itemize}

\textbf{Supervised and Unsupervised Warping}: In \emph{unsupervised warping}, the warping path is determined by minimizing a distance function, such as Mean Square Error (MSE) or Mean Absolute Error (MAE), to align signals without considering labels. This approach is used when signals have no labels or are aligned independently of them. In contrast, \emph{supervised warping} aligns signals with similar labels while distancing those with different labels.

\textbf{Linear and Nonlinear Warping}: In linear warping, represented as $Y(t)=X(at+b)$ with $a,b\in\mathbb{R}$, the warping path follows a linear function of time. However, in most practical cases, a more complex function is needed for accurate alignment. Nonlinear warping provides greater flexibility to better capture the relationships between signals.

\textbf{Warping Function and Warping Matrix}: The warped version of signal $X$ is denoted as $X_{warp}$, with the warping function $\tau(\cdot)$ representing the warping path, as expressed mathematically in Eq. \ref{eq1-2}.
\begin{equation}
\label{eq1-2}
X_{warp}(t)=X(\tau(t))
\end{equation}
For instance, in linear warping case, where $X_{warp}(t)=X(at+b)$, the warping function is $\tau(t)=at+b$. The warping matrix $W$ is defined such that $WX$ represents the warped form of $X$, allowing $X_{warp}$ to be represented in matrix form using Eq. \ref{eq1-3}.
\begin{equation}
\label{eq1-3}
X_{warp}=WX
\end{equation}

\subsection{DTW Problems}
DTW stands as the most widely method used for aligning time series. For brevity, we omit the introduction of DTW, and the reader is directed to \cite{dtw}. In this section, we address the challenges of DTW.

\textbf{Polynomial computational complexity}: The main limitation of DTW is its polynomial computational complexity, making it unsuitable for large datasets. To address this, various extensions have been developed to reduce the complexity from \emph{polynomial} to \emph{linear}. Speedup strategies fall into two categories: constraint addition and data abbreviation. In \cite{fdtw}, a linear-time algorithm is proposed, combining both approaches to offer a more efficient alternative to traditional DTW.

\textbf{Singularity}: A key issue in DTW is singularity, where differences in the vertical axis are misrepresented by warping the horizontal axis. This results in inconsistent alignments, with one point mapping to multiple points in another signal. To address this, it is crucial to consider the \emph{local shape} of the signal rather than just raw values. Solutions include using shape descriptors \cite{shape}, signal derivatives \cite{ddtw}, or employing a \emph{neural network} before warping to extract relevant features \cite{deep}, all of which help mitigate singularity and improve alignment accuracy.

\textbf{Non-differentiability}: A major limitation of DTW is its non-differentiability, making it challenging to use as a positive definite kernel or loss function in neural networks. To overcome this, researchers have developed approximate yet differentiable alternatives, such as Soft-DTW \cite{sdtw}.

\subsection{After DTW}
In an attempt to address the limitations of DTW, several alternative methods have been proposed:
\begin{itemize} \itemsep0em \large
\item \textbf{Generalized Time Warping (GTW) \cite{gtw}:} GTW addresses the polynomial complexity of DTW by introducing a linear-time algorithm that models the warping path as a linear combination of basis functions.
\item \textbf{Trainable Time Warping (TTW) \cite{ttw}:} TTW enhances warping by operating in the continuous time domain with convolutional kernels, offering better performance for complex warpings.
\item \textbf{Neural Time Warping (NTW) \cite{ntw}:} NTW relaxes the original DTW optimization problem to a continuous convex problem and finds the solution using a neural network.
\end{itemize}
Both TTW and NTW serve as approximations of the original DTW problem. Additionally, studies \cite{pcf} and \cite{ndm} introduce modifications to DTW to enhance its effectiveness in time series classification.

\subsection{Using Deep Learning}
Integrating deep neural networks, such as Convolutional Neural Networks (CNN) or Recurrent Neural Networks (RNN), into time series alignment provides significant advantages due to their structural flexibility, adaptable loss functions, and tunable hyperparameters. Their ability to extract meaningful features helps overcome challenges like the singularity problem in DTW.
\begin{itemize} \itemsep0em \large
\item \textbf{Supervised Warping with Deep Learning \cite{nwarp}:} This approach performs supervised warping using feature extractor and warper networks, generating a similarity index and a warping path for each time series pair. However, the warping path is a by-product, with no guarantee of its validity.
\item \textbf{Sequence Transformer Network (STN) \cite{stn}:} STN, built on CNN, enables simple translations and scalings in both time and amplitude domains. This provides a powerful deep learning-based tool for time series alignment.
\item \textbf{Temporal Transformer Network (TTN) \cite{ttn}:} TTN is a supervised warping module placed before a classifier to reduce intra-class variability and increase inter-class separation, improving classification performance.
\end{itemize}

\subsection{Contributions}
In our work, we have introduced the following contributions:
\begin{itemize} \itemsep0em \large
\item \textbf{Linear Computational Complexity:} Our model achieves linear inference complexity, addressing the polynomial complexity issue found in many previous MSA/MTSA methods.
\item \textbf{Grouped MTSA Algorithm:} Instead of performing multiple pairwise alignments like many previous MSA/MTSA methods, our proposed grouped MTSA algorithm enhances efficiency and scalability.
\item \textbf{Deep Neural Network Utilization:} By leveraging a deep neural network with an appropriate loss function, we address some drawbacks of DTW, improving the model’s ability to capture complex time series relationships.
\item \textbf{Decomposition of Nonlinear Warpings:} We break down complex nonlinear warpings into piecewise linear segments, enabling varying levels of complexity through simple linear warpings for a more flexible and adaptive approach.
\item \textbf{Warping Constraints Guarantee:} Our approach ensures compliance with the three warping constraints, maintaining proper chronological order and continuity in alignment.
\item \textbf{Improved Classification Accuracy:} Using our MTSA method before classification has led to increased accuracy across nearly all UCR Archive 2018 datasets.
\end{itemize}

\section{The Proposed Method} \label{sec3}
\subsection{MTSA Problem Definition}
Suppose $N$ time series $X_1,X_2,...,X_N$, where for $i\in[1:N]$, $X_i\in\mathbb{R}^{d_i\times T_i}$ with $d_i$ and $T_i$ representing the dimension and length of $X_i$, respectively. Two models can be employed to express time warping:
\begin{itemize} \itemsep0em \large
\item \textbf{Matrix Multiplication:} Defining warping matrices as $W_i$ for $i\in[1:N]$, the warped form of $X_i$ can be expressed as $W_iX_i$, as detailed in Section \ref{sec2-1}. One possible MSE cost function for the MTSA problem can be formulated as shown in Eq. \ref{eq4}:
\begin{equation}
\label{eq4}
J_{MTSA1}(\{W_i\})=\sum_{i=1}^N\sum_{j=1}^N ||W_iX_i-W_jX_j||^2_F
\end{equation}
\item \textbf{Function Composition:} Utilizing warping functions $\tau_i$ for $i\in[1:N]$, the warped form of $X_i$ is $X_i\circ\tau_i=X_i(\tau_i(t))$ and the associated cost function can be expressed as shown in Eq. \ref{eq5}:
\begin{equation}
\label{eq5}
J_{MTSA2}(\{\tau_i\})=\sum_{i=1}^N\sum_{j=1}^N ||X_i(\tau_i(t))-X_j(\tau_j(t))||^2_F
\end{equation}
\end{itemize}

\subsection{Warping Function and Constraints}
\begin{figure}
\captionsetup{justification=centering}
\begin{center}
\includegraphics[width=0.4\linewidth]{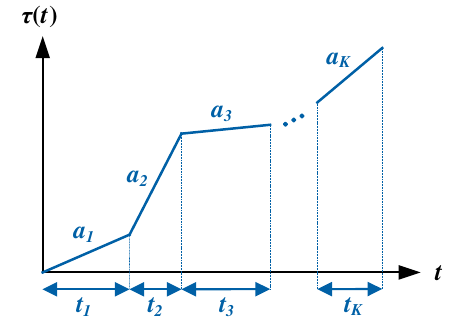}
\vspace*{2mm}
\caption{\label{fig1} The implemented warping function $\tau(t)$.}
\end{center}
\end{figure}

A linear warping function $\tau(t)=at+b$ can be implemented using a neural network with two output parameters ($a$ and $b$). However, this function is too simplistic for real-world scenarios. Instead, we adopt a more generalizable piece-wise linear function, as depicted in Fig. \ref{fig1}. It has slope $a_1$ in $t\in[0,t_1)$, $a_2$ in $t\in[t_1,t_1+t_2)$, ..., and $a_K$ in $t\in[\sum_{k=1}^{K-1}t_k,\sum_{k=1}^{K}t_k)$. Increasing K introduces more non-linearity into the model. In this case, the neural network must output $2K$ non-negative parameters: $\{a_1,a_2,...,a_K,t_1,t_2,...,t_K\}$. The mathematical formulation of the warping function $\tau(t)$ is given in Eq. \ref{eq6}.
\begin{equation}
\label{eq6}
\tau(t)=
\begin{cases}
a_1t &  t<t_1 \\
a_1t_1+a_2(t-t_1) &  t_1\leq t<t_1+t_2 \\
... & ...  \\
\sum_{k=1}^{K-1}a_kt_k+a_K(t-\sum_{k=1}^{K-1}t_k) & \sum_{k=1}^{K-1}t_k\leq t<\sum_{k=1}^{K}t_k
\end{cases}	
\end{equation}

\textbf{The warping constraints}:We verify the validity of the three warping constraints in the warping function shown in Fig \ref{fig1}. 
\begin{itemize} \itemsep0em \large
\item \emph{Boundary condition}: It is evident that $\tau(0)=0$. Additionally, we enforce $\sum_{k=1}^{K}t_k=T$, where $T$ is the length of the target warped signal.
\item \emph{Monotonicity condition}: This condition holds if $a_k\geq0$ for $k\in[1:K]$. Ensuring non-negative slopes guarantees a monotonically increasing warping function.
\item \emph{Continuity condition}: The function $\tau(t)$ is continuous, thus satisfying the continuity constraint.
\end{itemize}

\subsection{Non-differentiability Problem}
Consider a neural network is trained to implement the warping function $\tau(\cdot)$, and let signal $X$ with length $T$ be inputted to the network. The warped signal is obtained as $X_{warp}=X(\tau(\cdot))$. Consequently, $X(\tau(t))$ should be calculated for each $t\in[1,T]$. 

However, if $\tau(t)$ is not an integer, standard (\textit{hard}) warping approximates it to the nearest integer since $X$ is defined only at discrete time steps. This makes the loss function non-differentiable, as small changes in time ($t_k$) or amplitude ($a_k$) parameters may result in non-integer $\tau(t)$, causing $X(\tau(t))$ and the loss function to be undefined. Consequently, gradient-based optimization cannot be applied.

 To solve this, soft warping is introduced, allowing $\tau(t)$ to be a floating-point value. The warped signal $X_{warp}$ is then computed using interpolation. This interpolation is modeled through matrix multiplication (Eq. \ref{eq4}), where the warping matrix $W$ contains values in the range [0,1].

\subsection{Neural Network Structure}

\begin{figure*}[h!]
\begin{center}
\includegraphics[width=0.7\linewidth]{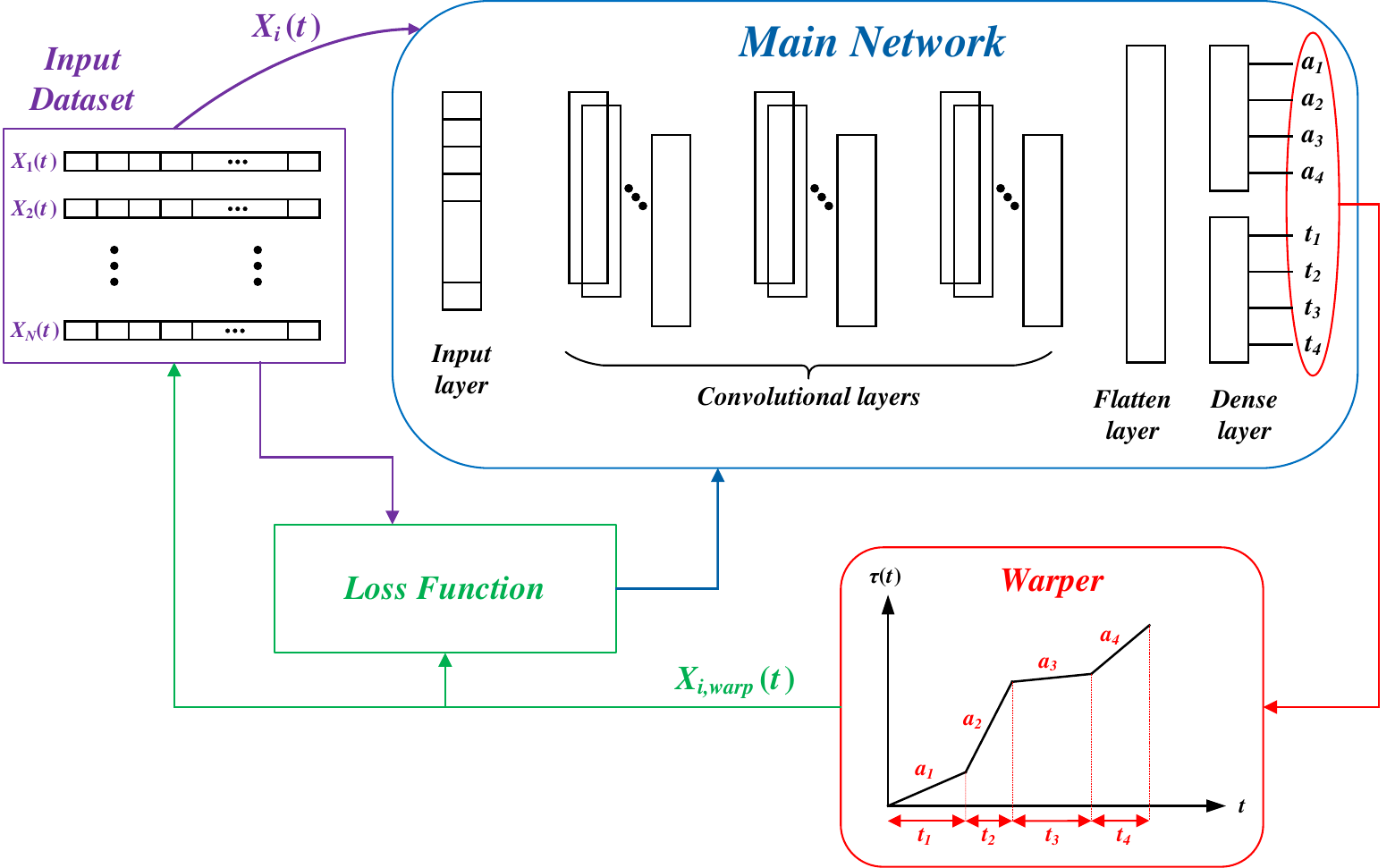}
\vspace*{2mm}
\caption{\label{fig2} The overall structure of the network.}
\end{center}
\end{figure*}

The overall structure of the neural network is illustrated in Fig. \ref{fig2}. The input time series $X_1(t),X_2(t),...,X_N(t)$ are assumed to have the same length at this stage; considerations for different-length time series will be addressed later. The primary network is a CNN with an input, three convolutional, a flatten and two dense layers.
\begin{itemize} \itemsep0em \large
\item \textbf{Input Layer:} Receives $X_i(t)$ from the dataset and passes it to the first convolutional layer.
\item \textbf{Convolutional Layers:} Comprise multiple convolutional kernels and pooling layers to extract features.
\item \textbf{Flatten Layer:} Converts the final convolutional layer’s output into a vector proportional to the input time series length.
\item \textbf{Parallel Dense Layers:} Two parallel dense layers generate the warping function parameters $\{a_1,a_2,a_3,a_4\}$ and $\{t_1,t_2,t_3,t_4\}$, as shown in Fig. \ref{fig1} for $K=4$.
\end{itemize}
From these outputs a warping function is implemented, and a warping matrix $W_i$ is calculated using the soft warping concept. The warped input $X_{i,warp}(t)$ is obtained by multiplying $X_i$ with $W_i$ and is applied to both the loss function and the input dataset blocks.

Two key contributions related to the neural network include the \textbf{loss function block} and the\textbf{training procedure}, which will be discussed in the following subsections.

\subsection{Loss Function}
As discussed in Section \ref{sec2}, DTW faces issues like computational complexity and singularity. To address \textit{singularity}, we propose two solutions: First, using convolutional kernels in CNNs for feature extraction, allowing local patterns at each temporal point to influence adjacent points, creating relationships between them. Second, instead of relying on traditional DTW algorithms with MSE loss functions, which can cause singularity due to their point-wise nature, we implement a more robust loss function that captures the overall similarity between two signals, rather than just point-to-point proximity.

The time warping loss function must accommodate small to moderate scalings and shifts in the temporal domain without correcting amplitude. So, when two signals are multiples of each other, the loss function should reach its minimum. The approach is to apply the \textit{inner product} of the two signals. For two arbitrary 1-dimensional signals $X$ and $Y$ (vectors), the \textit{Cosine Similarity} function is defined as follows:

\begin{equation}
\label{eq8}
S_C(X,Y)=\frac{<X,Y>}{\max\{||X||_2\;||Y||_2,\epsilon\}}
\end{equation}
Here, $||\cdot||_2$ denotes the Euclidean norm, and $\epsilon$ is a small positive constant to prevent division by zero. Cosine similarity ranges from $[-1,1]$, where 1 signifies codirectional signals, 0 indicates orthogonal signals, and -1 represents contradirectional signals. To achieve smoother results, we use a quadratic form of cosine similarity while preserving its sign. This is because both orthogonal and contradirectional signals are undesirable, and we need codirectional signals. Consequently, the loss function in Eq. \ref{eq9} is defined using the signed square form of cosine similarity.

\begin{equation}
\label{eq9}
L(X,Y)=1-S_C(X,Y)^2\mathrm{sign}(S_C)
\end{equation}
Finally, the main loss function between two arbitrary signals $X$ and $Y$ is introduced as Eq. \ref{eq9-2}:
\begin{equation}
\label{eq9-2}
L_{main}(X,Y)=L(X_{warp},Y)
\end{equation}
The main loss function in Eq. \ref{eq9-2} is similar to Eq. \ref{eq9}, only the first signal ($X$) is warped and then its cosine similarity with the second signal ($Y$) is measured.

If the signals are matrices (i.e., dimensions greater than one), each row is treated as an individual vector. Cosine similarity is then calculated between corresponding rows using Eq. \ref{eq8}. This results in a vector as the main loss function in Eq. \ref{eq9-2}, with a size equal to the signal dimensions. To obtain a specific loss function, the average value of the elements in this vector is computed.

\begin{figure}
\begin{center}
\includegraphics[width=0.45\linewidth]{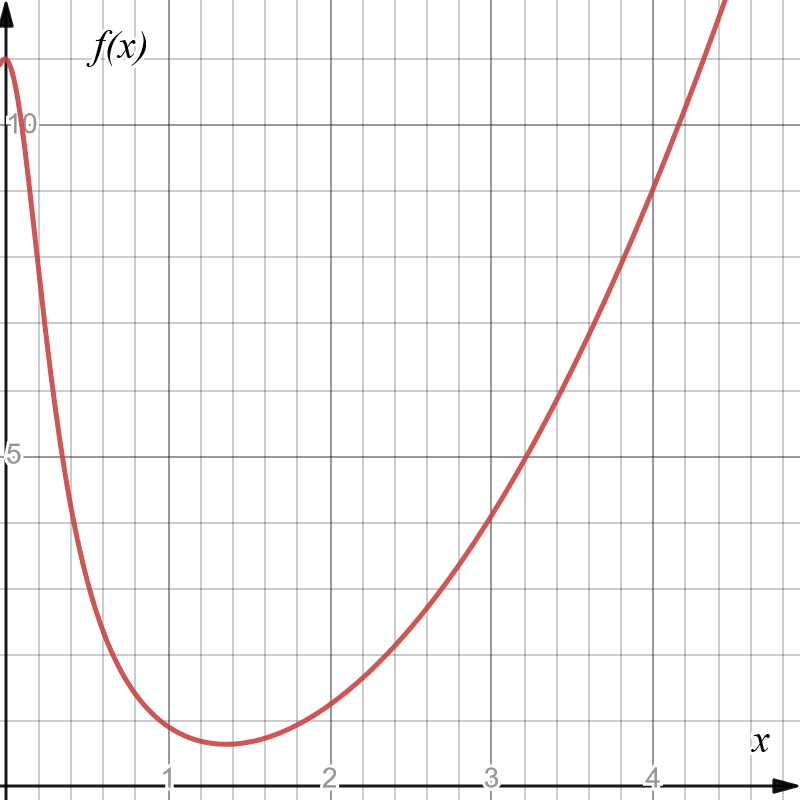}
\vspace*{2mm}
\caption{\label{fig3} A graphical curve from the prototype penalization function.}
\end{center}
\end{figure}

In the implemented warping function (see Fig. \ref{fig1}), it is evident that $t_i\geq0$ for $i\in[1:K]$. During training, we enforce $\sum_{k=1}^{K}t_k=T$, where $T$ is the time series length. Satisfying the monotonicity condition requires $a_i\geq0$ for $i\in[1:K]$. If $a_i=1$ for all $i\in[1:K]$, the warping function becomes the identity, implying no change to the signal. Since signals in the dataset are assumed to be homogeneous with minimal discrepancies, the values of $\{a_1,...,a_K\}$ should stay close to 1. To encourage this, two \emph{penalization terms} are added to the loss function. Suppose $x$ is a measure of the mean amplitude of $\{a_1,...,a_K\}$. We define two functions on $x$:
\begin{itemize} \itemsep0em \large
\item $f_1(x)=(x-1)^2$: Encourages $x$ to be around 1 and penalizes $x$ for values far larger than 1.
\item $f_2(x)=1/(x^2+\epsilon)$: Prevents $x$ from going too close to zero. Here, $\epsilon$ is a small positive constant.
\end{itemize}
The combination of these two functions can be expressed as Eq. \ref{eq10}, and Fig. \ref{fig3} illustrates its graphical curve.
\begin{equation}
\label{eq10}
f(x)=(x-1)^2+\frac{1}{x^2+0.1}
\end{equation}
Based on Fig. \ref{fig3}, the function in Eq. \ref{eq10} can serve as an effective penalization term. Building on this prototype, we define the following penalization function:
\begin{equation}
\label{eq11}
L_{pen.}(a_1,...,a_K)=\sum_{k=1}^{K}(a_k-1)^2+\lambda_1\frac{1}{\frac{1}{K}\sum_{k=1}^{K}a_k^2+0.1}
\end{equation}
Finally, combining Eq. \ref{eq11} with Eq. \ref{eq9}, the loss function for an input time series $X$ can be expressed as Eq. \ref{eq12}:
\begin{equation}
\label{eq12}
\begin{aligned}
& L_{final}(X,Y)=L_{main}(X,Y)+\lambda_2L_{pen.}(a_1,...,a_K) \\ 
& =1-S_C(X_{warp},Y)^2\mathrm{sign}(S_C) \\ 
& +\lambda_2\left(\sum_{k=1}^{K}(a_k-1)^2+\lambda_1\frac{1}{\frac{1}{K}\sum_{k=1}^{K}a_k^2+0.1}\right)
\end{aligned}
\end{equation}
In Eqs. \ref{eq11} and \ref{eq12}, $\lambda_1$ and $\lambda_2$ are hyper-parameters that control the strength of the penalization terms, while $a_k$ for $k\in[1:K]$ are the amplitude outputs of the network corresponding to the input $X$. The main loss function, $L_{main}(X,Y)$, is computed between the warped form of the input signal $X_{warp}$ and the second signal $Y$. For two signals $X$ and $Y$, the neural network can warp the first signal $X$ to align with $Y$ using Eq. \ref{eq12}. For more than two time series, the problem becomes MTSA, which will be discussed in the next subsection.

\subsection{Training and Testing Procedure}
In this section, we explain how our framework extends to the multiple time series case for the MTSA problem. Consider Fig. \ref{fig2}, where the signals in the input dataset $X_i$ for $i\in[1:N]$ have the same length $T$. If their lengths differ, a pre-processing stage will equalize them. Below is the proposed algorithm for the training procedure:
\begin{enumerate} \large
\item Apply each time series $X_i$ to the network input.
\item Obtain amplitude parameters $\{a_1,a_2,a_3,a_4\}$ and time parameters $\{t_1,t_2,t_3,t_4\}$ from the network.
\item Utilize the warper block to generate the warping matrix associated with these values and multiply it with the input time series to construct $X_{i,warp}$.
\item The loss function block calculates the average final loss between $X_{i,warp}$ and each of the other $N-1$ signals according to Eq. \ref{eq12}.
\item Replace the original $X_i$ with its warped version $X_{i,warp}$.
\item Repeat steps 1-5 for all $N$ signals, completing one epoch of training.
\item Perform an appropriate number of epochs to gradually align signals to each other.
\end{enumerate}

Substituting signals with their warped versions is essential in our MTSA framework. However, early in training, the network may lack meaningful warpings. Delaying substitution until the model learns more relevant information ensures stable and informed dataset updates.

Ultimately, the network aligns $N$ input signals, enabling accurate warping of homogeneous test time series. During \textit{testing} (illustrated in Fig. \ref{fig2}), the process remains the same except for omitting the loss function block. The input test signal $X_i$ is processed by the network, producing the warped test signal $X_{i,warp}$, via the warper block.

A key benefit of using deep neural networks for time series alignment is the elimination of backpropagation during testing. Unlike conventional methods such as DTW, which require repeated optimization for each alignment, our approach uses a parameterized network that learns to align signals efficiently.

\section{Experiments}
This paper conducts four experiments using the UCR Time Series Classification Archive \cite{ucr}, which includes 128 univariate time series datasets. The first experiment addresses the MTSA problem by aligning test signals to training signals. The second experiment explores warped averaging as a key MTSA application, highlighting notable cases to evaluate the method's performance. The third experiment involves a classification test on 90 datasets, reporting accuracy for a Nearest Neighbor classifier in four scenarios: no warping, DTW, DBA, and the proposed approach. The fourth experiment validates the method's superiority by measuring classification rate and error using a deep ResNet classifier.

The convolutional neural network consists of three layers with filter sizes of 13, 7, and 3, and filter counts of 128, 64, and 32, respectively. Each convolutional layer is followed by an average pooling layer (stride 1, sizes 6, 4, and 2). After the third layer, the tensor is flattened and processed by two \textit{parallel} dense layers, each with 4 output neurons representing $\{a_1,a_2,a_3,a_4\}$ and $\{t_1,t_2,t_3,t_4\}$. ReLU activation ensures non-negative, unbounded outputs for $a$ and $t$.

The hyperparameters $\lambda_1$ and $\lambda_2$ in Eq. \ref{eq12} are set to 0.5 for most datasets. Although optimizing them individually could improve results, we avoided this due to its time-intensive nature. The learning rate is fixed at $10^{-3}$. Training runs for 25 epochs, with checkpoints saved every 5 epochs to account for potential early stopping benefits. The best model is chosen based on validation accuracy. The implementation uses the PyTorch library.

\subsection{The Multiple Time Series Alignment (MTSA)}

A key application of MTSA is computing a \textit{warped average} to represent a set of signals, as a simple arithmetic average cannot handle temporal shifts or scale variations. DBA \cite{dba}, a robust MTSA method, iteratively uses DTW to align signals with an evolving average. In this study, DBA is used as the baseline for MTSA (in this section) and warped averaging (in the next section) to demonstrate the advantages of our proposed time series alignment approach.

\begin{figure*}
\begin{center}
\includegraphics[width=0.9\linewidth]{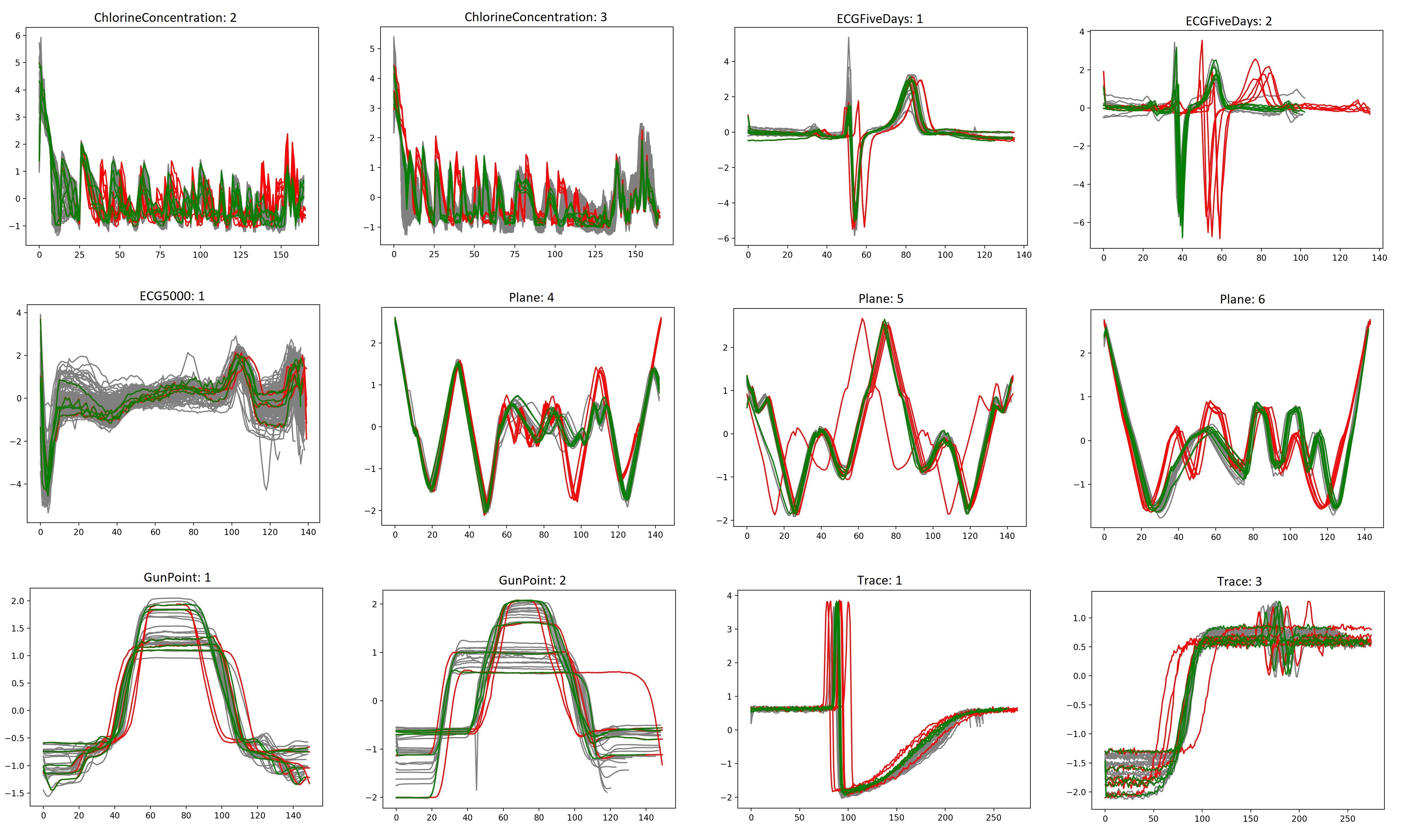}
\caption{\label{fig3-2} Results of the MTSA experiment, with dataset names and labels displayed above each. In each plot, gray signals represent the warped training signals, while red signals indicate five randomly selected test signals requiring alignment. The green signals show the warped versions of the red signals, generated by our model.}
\end{center}
\end{figure*}

For each dataset, signals with the same label are inputted into the model to ensure homogeneity. Standard UCR dataset train-test splits are used, with the training set for model training. The goal is to optimally align five test signals with their corresponding training signals. Fig. \ref{fig3-2} illustrates results for various datasets and labels, showing red signals warped to align with gray signals, producing green signals. In cases like \textit{"Plane: 4"} and \textit{"Trace: 3"}, simple linear transformations are insufficient, requiring more complex non-linear warpings for accurate alignment.

For each test signal (red), generating its warped counterpart (green) involves solving an MTSA problem to align it with a set of training signals (gray). Once the warper network is trained, the MTSA problem is solved by passing the test signal through the network, ensuring linear computational complexity relative to signal length. Notably, inference time is unaffected by the number of training signals, making the method scalable for large datasets. A major advantage of deep neural networks is the decoupling of training time (a one-time process) from test time.

\begin{figure}
\begin{center}
\includegraphics[width=0.5\linewidth]{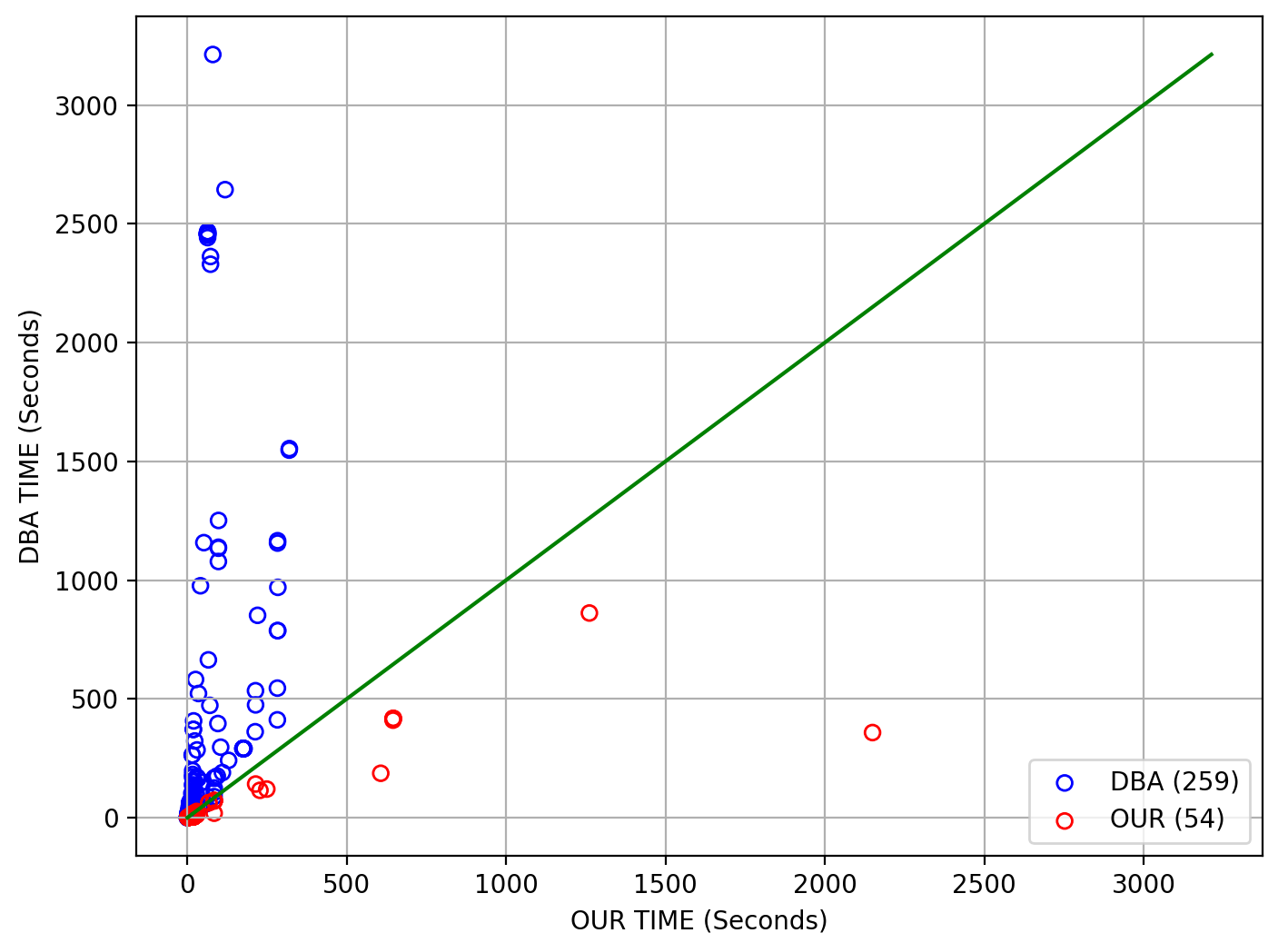}
\caption{\label{fig3-4} Scatter plot comparing the timing of our method with DBA. Each point represents a label of a dataset, with points above the $y=x$ line indicating a win for our model (blue points - our time is less than dba time) and those below showing a loss (red points - our time is more than dba time).}
\end{center}
\end{figure}

For comparison, we assess the computation time of DBA \cite{dba} for generating warped averages of signals, followed by DTW to align each test signal with the training set. While the quality of the warped average is discussed in Subsection \ref{sec4}, this section focuses on timing results. As shown in Table \ref{tab0}, our model’s total processing time is, on average, more than twice faster than the DBA-based method. Figure \ref{fig3-4} provides a detailed comparison across all UCR datasets, showing that our model is faster in over 82\% of cases. Notably, it reduces DBA’s computation time from 258 to 59 seconds, achieving more than a 4-fold improvement.

\begin{table}
\begin{center}
\begin{minipage}{\linewidth}
\fontsize{7pt}{7pt}\selectfont
\caption{Timing comparison for an MTSA problem between our approach and a DBA-based approach.} \label{tab0}
\begin{tabular}{||c c c| c c c c||}
\hline
\textbf{\makecell{Dataset\\name}} & \textbf{Label} & \textbf{\makecell{\# of \\ Train \\ signals}} & \textbf{\makecell{OUR time: \\ Train (sec)}} & \textbf{\makecell{OUR Time : \\ Test (sec)}} & \textbf{\makecell{OUR Time : \\ Whole (sec)}} & \textbf{\makecell{DBA Time : \\ Whole (sec)}} \\ [0.5ex] 
\hline
\hline
\makecell{Chlorine\\Concentration} & 2 & 91 & 11.6 & 2.27 & 13.87 & 87.7 \\
\hline
\makecell{Chlorine\\Concentration} & 3 & 262 & 102.4 & 2.24& 104.6 & 259.9 \\
\hline
ECG5000 & 1 & 292 & 127.6 & 1.65 & 129.2 & 201.6 \\
\hline
ECGFiveDays & 1 & 14 & 0.30 & 1.51 & 1.81 & 3.94 \\
\hline
ECGFiveDays & 2 & 9 & 0.13 & 1.55 & 1.68 & 2.04 \\
\hline
GunPoint & 1 & 24 & 0.81 & 1.89 & 2.7 & 11.4 \\
\hline
GunPoint & 2 & 26 & 0.95 & 2.02 & 2.97 & 12.9 \\
\hline
Plane & 4 & 16 & 0.38 & 1.73 & 2.11 & 5.41 \\
\hline
Plane & 5 & 13 & 0.25 & 1.71 & 1.96 & 3.97 \\
\hline
Plane & 6 & 18 & 0.46 & 1.77 & 2.23 & 6.56 \\
\hline
Trace & 1 & 26 & 0.96 & 6.15 & 7.11 & 42.3 \\
\hline
Trace & 3 & 22 & 0.70 & 6.15 & 6.85 & 32.4 \\
\hline
\hline
\end{tabular}
\end{minipage}
\end{center}
\end{table}

\subsection{Representative and Warped Averaging}\label{sec4}
In this section, we provide visual comparisons demonstrating the advantages of our approach over the DBA algorithm in computing the warped average signal and effectively addressing various challenges.

\begin{figure}
\begin{center}
\includegraphics[width=1\linewidth]{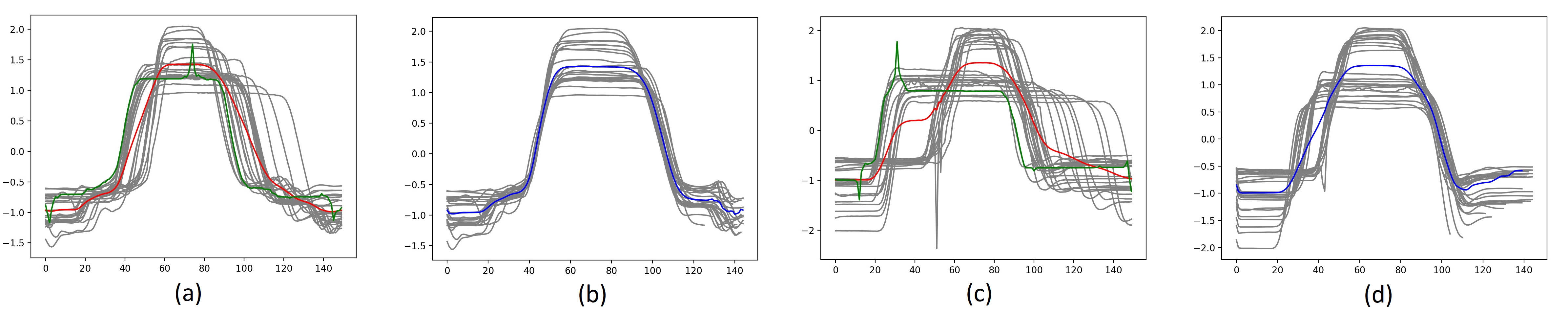}
\caption{\label{fig4} Results on the GunPoint dataset. (a) label 1, gray: original time series, red: simple average, green: DBA signal. (b) label 1, gray: warped time series with our method, blue: warped average. (c) label 2, gray: original time series, red: simple average, green: DBA signal. (d) label 2, gray: warped time series with our method, blue: warped average.}
\end{center}
\end{figure}

\textbf{Overall Comparison:} An overall test on the GunPoint dataset evaluates our method's performance, as shown in Fig. \ref{fig4}. Fig. \ref{fig4} (a) displays label 1 signals with a simple average (red) and DBA signal (green). Fig. \ref{fig4} (b) shows warped signals using our method and their average (green). Fig. \ref{fig4} (c) and \ref{fig4} (d) present the same for label 2. The results highlight that the simple average fails to capture slightly complex trends, particularly for label 2, while DBA introduces unwanted spikes. In contrast, our method aligns signals effectively, producing a warped average that preserves the trend of its signals and serves as a representative for each class.

\begin{figure}
\begin{center}
\includegraphics[width=0.5\linewidth]{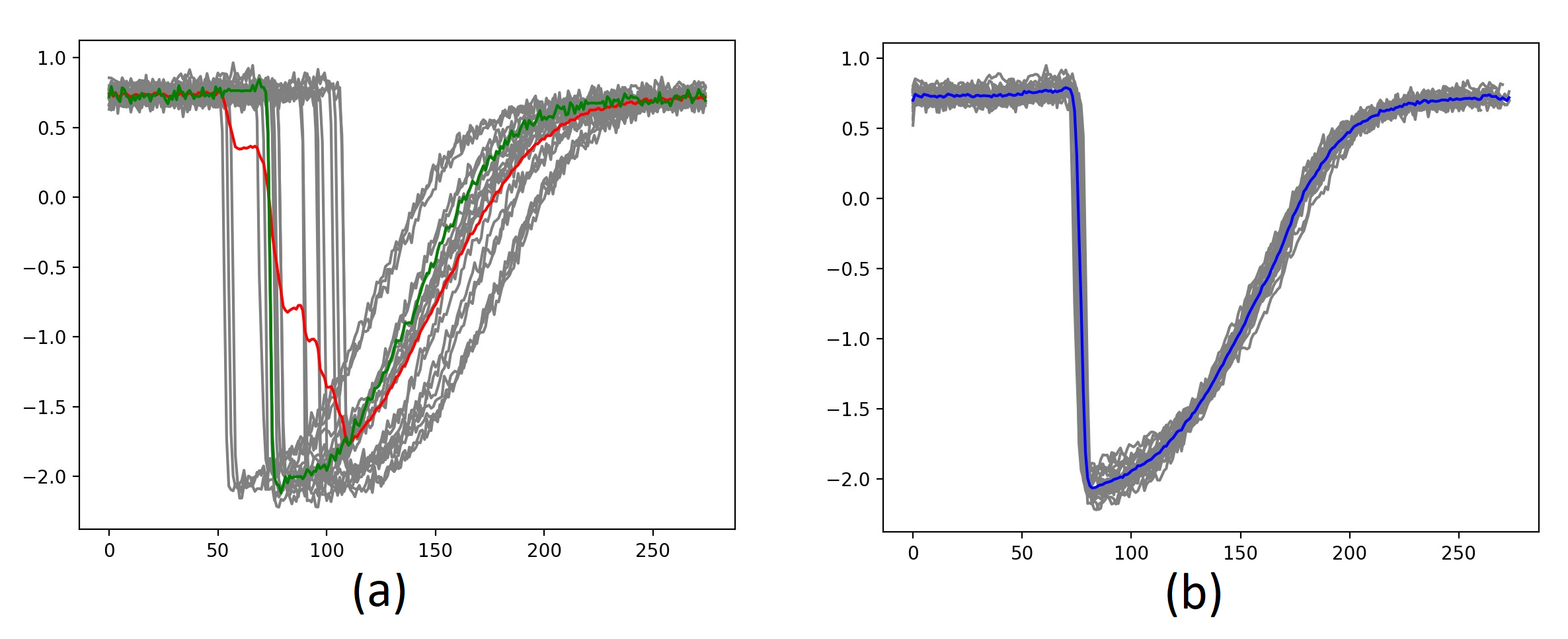}
\caption{\label{fig5} Results on the Trace dataset, label 2. For details refer to Fig. \ref{fig4} caption.}
\end{center}
\end{figure}

\textbf{Preserve Signal Shapes:} Preserving signal shapes is crucial in warped averaging, especially for challenging datasets like Trace. Simple averaging fails to capture the true shape of signals, as shown in Fig. \ref{fig5}(a). While DBA improves the results, our approach, illustrated in Fig. \ref{fig5}(b), effectively compensates for signal shifts by applying appropriate multiple warping. This generates a warped average with reduced variations and better representation of the underlying trend compared to DBA.

\begin{figure}
\begin{center}
\includegraphics[width=1\linewidth]{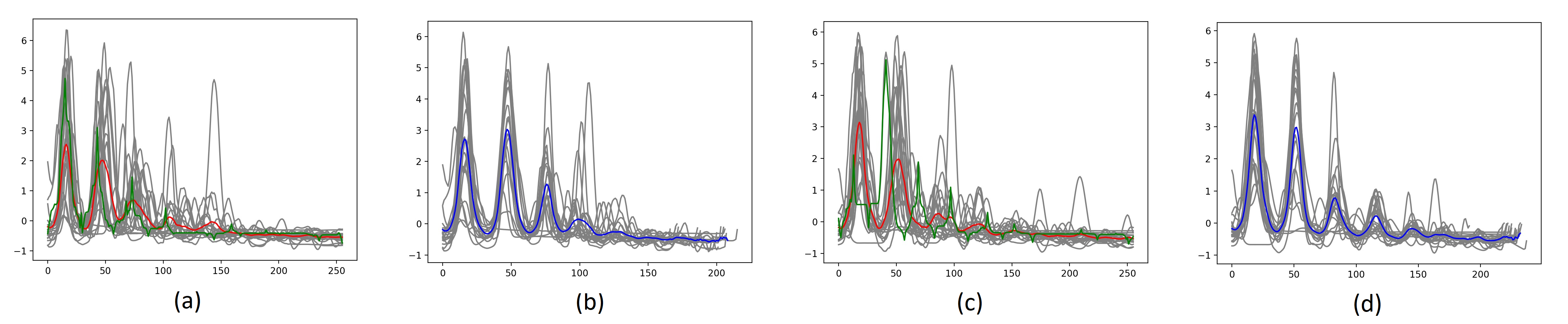}
\caption{\label{fig6} Results on the InsectWingbeatSound dataset, (a), (b): label 2 and (c), (d): label 10. For details refer to Fig. \ref{fig4} caption.}
\end{center}
\end{figure}

\textbf{Alignment of Peaks:} The InsectWingbeatSound dataset contains signals with sequences of unaligned peaks, making alignment and trend extraction very challenging. Fig. \ref{fig6}(a),(c) demonstrate that both simple averaging and DBA fail to preserve the sequence of peaks, particularly smaller ones. In contrast, Fig. \ref{fig6}(b),(d) show that warped signals and their averages successfully maintain the peak sequences.

\begin{figure}
\begin{center}
\includegraphics[width=1\linewidth]{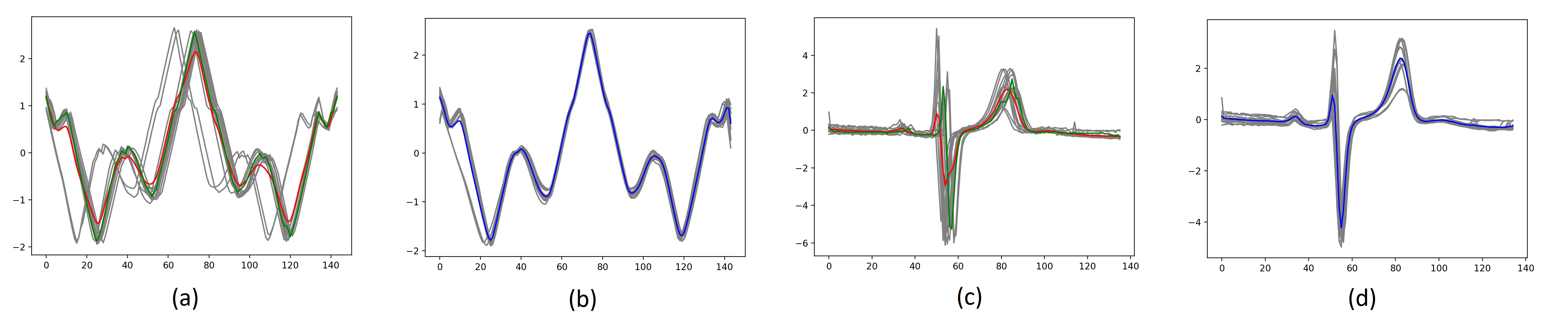}
\caption{\label{fig7} (a), (b): Results on the Plane dataset, label 5. (c), (d): Results on the ECGFiveDays dataset, label 1. For details refer to Fig. \ref{fig4} caption.}
\end{center}
\end{figure}

\textbf{Signal Shifts:} Time warping effectively compensates for temporal shifts in signals with similar shapes. As demonstrated in Fig. \ref{fig7}, our method successfully removes temporal displacements, resulting in warped signals that produce a more accurate average trend compared to other approaches.

\begin{figure}[t]
\begin{center}
\includegraphics[width=1\linewidth]{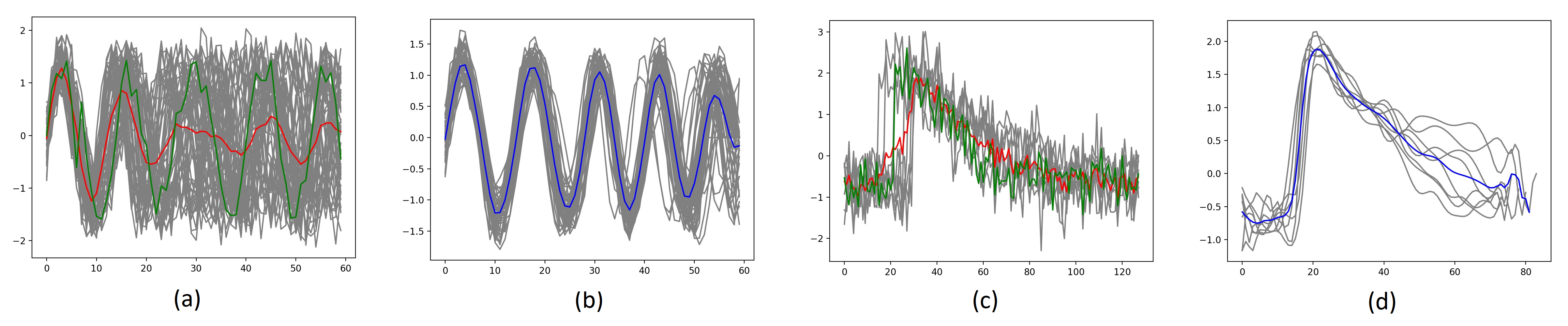}
\caption{\label{fig8} (a), (b): Results on the SyntheticControl dataset, label 2. (c), (d): Results on the CBF dataset, label 3. For details refer to Fig. \ref{fig4} caption.}
\end{center}
\end{figure}

\textbf{Noisy Environments:} Extracting signal shapes from datasets with high variation and noise is challenging. However, as shown in Fig. \ref{fig8} on the SyntheticControl and CBF datasets, our method effectively aligns signals and extracts a meaningful representative for the time series set, even under noisy conditions.

\begin{figure}
\begin{center}
\includegraphics[width=0.5\linewidth]{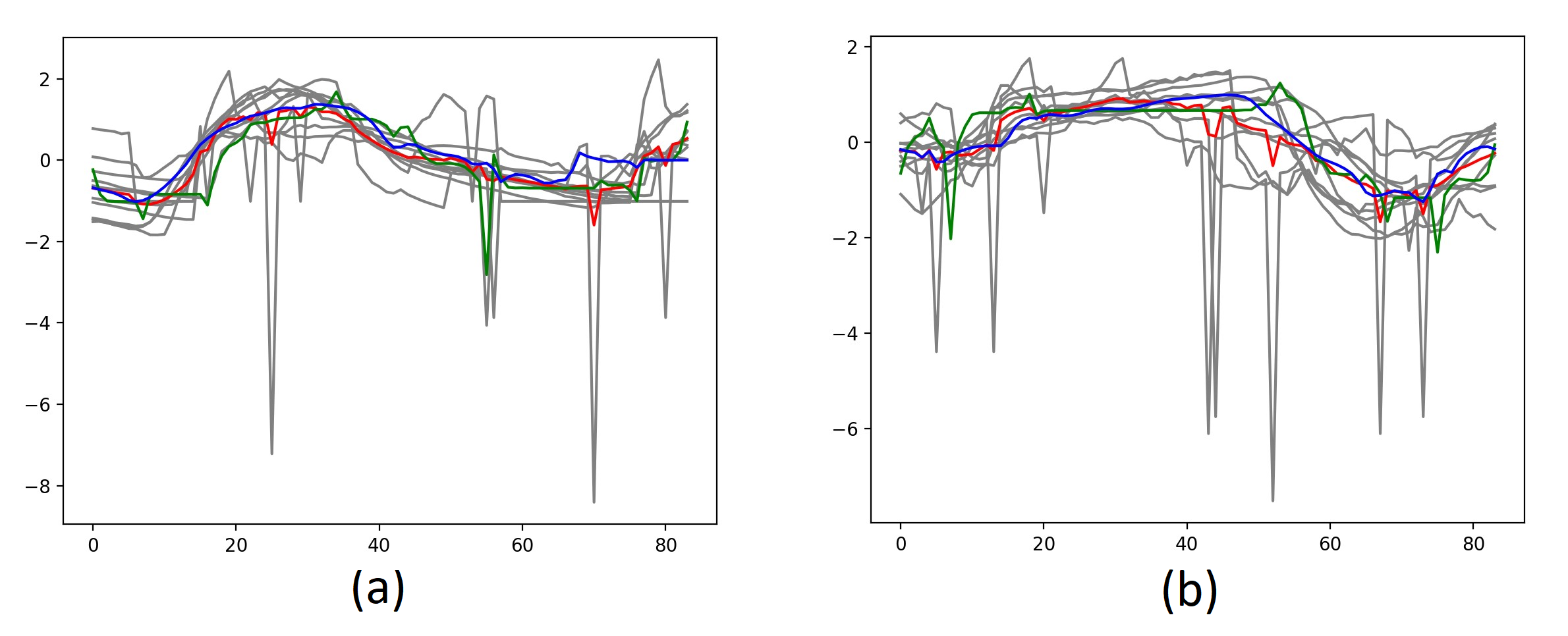}
\caption{\label{fig9} Results on the MoteStrain dataset. gray: original time series, red: simple average, green: DBA signal, blue: warped average with our method. (a): label 1,  (b): label 2.}
\end{center}
\end{figure}

\textbf{Outlier Signals:} If rare signals exhibit peaks around a specific temporal point, these should likely be interpreted as outlier trends and excluded from the representative signal. As demonstrated in Fig. \ref{fig9}, which presents results on the MoteStrain dataset, local peaks are reflected in both the average and DBA signals. However, averaging from warped signals with our model gives a representative signal that captures the overall trend without the local peaks. 

\subsection{The Comprehensive Classification Test}
This section and the next aim to show how our proposed warper network enhances classification quality, using \textit{classification accuracy} as the metric. Since classification is not the main focus, we employ the simplest classifier, nearest neighbor (NN), and evaluate accuracy across datasets under four conditions: a basic NN classifier, and NN combined with DTW, DBA, and our method.

In the DTW+NN classifier, the Euclidean distance is replaced with DTW distance, requiring DTW computation between the test sample and all training signals. In the DBA approach, the warped average of training signals is computed for each class, and test samples are assigned to the class whose representative has the smallest DTW distance.

In our approach, a neural network is trained for each class using specified parameters. Training is repeated with multiple random initializations, and the best model is selected based on validation accuracy. The final model's performance is evaluated on the test dataset.

The UCR Archive contains 11 datasets of varying lengths, requiring a pre-processing step to equalize their lengths before inputting them into the network. Following \cite{akyash}, we compute the average series length and adjust each time series accordingly. For longer series, random time steps are removed, while for shorter ones, new points are inserted using the average of random time steps and their adjacent values. This method preserves the time series shape and is computationally more efficient than uniformly stretching the series, which would require recalculating all signal values.

After training on a dataset, each test signal is processed through all class-specific warpers. The error is measured between the warped test signal and \textit{the average of} all warped training signals for each class (warped by their corresponding class warper) using Eq. \ref{eq9}. The test signal is assigned to the class whose warper produces the smallest error.

A limitation of our approach is the requirement to train as many models as there are classes in a dataset, making it less practical for datasets with numerous classes. Due to this and resource constraints, we performed classification tests on 90 UCR Archive datasets. Table \ref{tab1} demonstrates that our method on average improves baseline results by \textbf{6.1\%}, DTW+NN by \textbf{3.1\%} and DBA+NN by \textbf{7.5\%}. The DBA approach yields the lowest accuracy because it compares test signals only to class representatives rather than all training signals (as in the Base and DTW methods). Additionally, using the same hyperparameters for most datasets resulted in slight accuracy reductions in some cases. We anticipate that fine-tuning will enhance these results.

The last row in Table \ref{tab1} shows the Mean Per Class Error (MPCE) introduced by \cite{resnet}, which is defined as Eq. \ref{eq13}.
\begin{equation}
\label{eq13}
MPCE=\frac{1}{K}\sum_{k=1}^{K}\frac{1-Acc_k}{Number\ of\ classes}
\end{equation}
In Eq. \ref{eq13}, $Acc_k$ is the classification accuracy in the $k$th dataset, and $K$ is the number of datasets. MPCE measures the expected error rate per class across all datasets. According to Table \ref{tab1}, our method reduces the MPCE by \textbf{24.6\%} compared to NN (0.0832 to 0.0627), \textbf{17.5\%} compared to DTW+NN (0.0760 to 0.0627) and \textbf{28.8\%} compared to DBA+NN (0.0881 to 0.0627). Thus, on average, it exhibits better classification accuracy per class for these 90 datasets.

The final column of Table \ref{tab1} shows the cosine similarity-based loss between training signals before and after the training process. Since some UCR datasets are manually aligned, applying a warper may not always enhance alignment. This can be observed by comparing the loss values of original and warped training signals. The datasets in Table \ref{tab1} are sorted by the degree of loss reduction after applying the network. Datasets in the top rows, which show greater loss reduction, also exhibit more significant accuracy improvements with our approach compared to the nearest neighbor (NN) method. In contrast, datasets in the lower rows (like OliveOil, Fungi, and Meat) are already well-aligned, so warping does not produce noticeable effects.

\begin{table}
\begin{minipage}{\linewidth}
\raggedright
\fontsize{7pt}{7pt}\selectfont

\caption{Classification accuracy comparison between our method and two base models over 90 datasets of the UCR Archive.} \label{tab1}
\hskip-0.51cm
\begin{tabular}{|c| c c c c| c|}
\hline
\textbf{Dataset name} & 
\textbf{\makecell{Base \\ NN}} & \textbf{\makecell{DTW \\ +NN}} & \textbf{\makecell{DBA \\ +NN}} & \textbf{OUR} & \textbf{\makecell{CS \\ org.}---\textgreater\makecell{CS \\ warp}}   \\ [0.5ex] 
\hline\hline 
ACSF1 & 54 & \textbf{64} & 47 & 60 & 0.326 --\textgreater 0.022 \\ 
\hline
Trace & 76 & \textbf{100} & 86 & 80 & 0.392 --\textgreater 0.042 \\
\hline
CBF & 85.5 & 71.7 & \textbf{92.2} & 90 & 0.745 --\textgreater 0.141 \\
\hline
TwoLeadECG & 78.5 & \textbf{93.1} & 87.1 & 90.5 & 0.247 --\textgreater 0.048 \\
\hline
SmoothSubspace & 95.3 & 81.3 & 82.7 & \textbf{96.7} & 0.275 --\textgreater 0.078 \\
\hline
ECG200 & 88 & \textbf{92.5} & 83 & 85 & 0.275 --\textgreater 0.085 \\
\hline
SonyAIBORobotSurface2 & \textbf{88.5} & 85.4 & 76.4 & 83 & 0.633 --\textgreater 0.218 \\
\hline
BME & 82.7 & 75 & 75.3 & \textbf{91.3} & 0.268 --\textgreater 0.120 \\
\hline
Car & 60 & \textbf{99.8} & 63.3 & 80 & 0.148 --\textgreater 0.068 \\
\hline
GunPoint & \textbf{91.3} & 88.7 & 76.7 & 87.7 & 0.366 --\textgreater 0.167 \\
\hline
Computers & 57 & \textbf{71.6} & 56.8 & 59.2 & 0.955 --\textgreater 0.452 \\
\hline
InlineSkate & 33.5 & 37.8 & 31.6 & \textbf{45.5} & 0.591 --\textgreater 0.285 \\
\hline
Plane & 96.2 & \textbf{100} & 99 & 99 & 0.101 --\textgreater 0.050 \\
\hline
AllGestureWiimoteZ & 47 & \textbf{65.4} & 53 & 54 & 0.634 --\textgreater 0.330 \\
\hline
PhalangesOutlinesCorrect & 77.5 & \textbf{93.1} & 75.9 & 90 & 0.078 --\textgreater 0.040 \\
\hline
UMD & 80.6 & \textbf{86.8} & 71.8 & 79.2 & 0.299 --\textgreater 0.161 \\
\hline
GunPointAgeSpan & 96 & \textbf{98.4} & 87.7 & 97 & 0.046 --\textgreater 0.025 \\
\hline
ECGFiveDays & 80 & 46.1 & 68.4 & \textbf{92.5} & 0.667 --\textgreater 0.364 \\
\hline
Fish & 78.3 & 81.1 & 69.1 & \textbf{81.5} & 0.087 --\textgreater 0.049 \\
\hline
Chinatown & \textbf{95} & \textbf{95} & 85.4 & \textbf{95} & 0.371 --\textgreater 0.215 \\
\hline
InsectWingbeatSound & \textbf{61} & 35.9 & 40.9 & 55.7 & 0.657 --\textgreater 0.403 \\
\hline
FreezerRegularTrain & 79 & \textbf{89.7} & 77.1 & 82.5 & 0.346 --\textgreater 0.223 \\
\hline
Yoga & 82 & 83.6 & 81.2 & \textbf{86.5} & 0.675 --\textgreater 0.445 \\
\hline
WormsTwoClass & 61 & 58.4 & 54.5 & \textbf{63.5} & 0.918 --\textgreater 0.609 \\
\hline
\tiny{ProximalPhalanxOutlineCorrect} & 77.5 & 78.3 & 74.6 & \textbf{80} & 0.034 --\textgreater 0.023 \\
\hline
SyntheticControl & 88.5 & 99 & 92.3 & \textbf{100} & 0.655 --\textgreater 0.446 \\
\hline
MedicalImages & 70.5 & 73.5 & 71.2 & \textbf{83} & 0.565 --\textgreater 0.388 \\
\hline
FreezerSmallTrain & 64.5 & 75.9 & 75.8 & \textbf{83} & 0.290 --\textgreater 0.200 \\
\hline
Meat & 93.3 & 93.3 & 90 & \textbf{100} & 0.000 --\textgreater 0.000 \\
\hline
Herring & 51.6 & 54.7 & 59.4 & \textbf{70.3} & 0.090 --\textgreater 0.064 \\
\hline
Lightning7 & 57.5 & 69.9 & 68.5 & \textbf{72.5} & 0.722 --\textgreater 0.512 \\
\hline
\tiny{MiddlePhalanxOutlineCorrect} & \textbf{76.5} & 71.1 & 68.1 & 69 & 0.050 --\textgreater 0.035 \\
\hline
ECG5000 & 91.5 & 75.6 & 84.5 & \textbf{95.5} & 0.374 --\textgreater 0.270 \\
\hline
FaceAll & 68 & \textbf{85.8} & 68.7 & 84.5 & 0.780 --\textgreater 0.568 \\
\hline
BirdChicken & 55 & 65 & 65 & \textbf{85} & 0.682 --\textgreater 0.496 \\
\hline
Wafer & \textbf{100} & 97.9 & 92.5 & 99.5 & 0.518 --\textgreater 0.382 \\
\hline
Symbols & 93.5 & \textbf{95.2} & 93.8 & 93.5 & 0.212 --\textgreater 0.158 \\
\hline
Worms & 45.5 & \textbf{61} & 45.5 & 59 & 0.899 --\textgreater 0.672 \\
\hline
ItalyPowerDemand & 97 & 95 & 92.7 & \textbf{98.5} & 0.312 --\textgreater 0.240 \\
\hline
\tiny{MiddlePhalanxOutlineAgeGroup} & 51.9 & 50.6 & 57.1 & \textbf{68.2} & 0.030 --\textgreater 0.023 \\
\hline
MiddlePhalanxTW & 51.3 & 50.6 & 48.7 & \textbf{62.3} & 0.018 --\textgreater 0.014 \\
\hline
\tiny{ProximalPhalanxOutlineAgeGroup} & 78 & 81 & 81.5 & \textbf{86} & 0.021 --\textgreater 0.017 \\
\hline
DistalPhalanxTW & 63.3 & 60.4 & 63.3 & \textbf{66.9} & 0.025 --\textgreater 0.020 \\
\hline
ArrowHead & \textbf{80} & 70.9 & 67.1 & \textbf{80} & 0.134 --\textgreater 0.109 \\
\hline
FordA & \textbf{68.5} & 56.8 & 62.5 & 63.5 & 0.473 --\textgreater 0.389 \\
\hline
FordB & 58 & 61.7 & 61.1 & \textbf{63} & 0.481 --\textgreater 0.396 \\
\hline
DiatomSizeReduction & 91.5 & 96.1 & 84.3 & \textbf{98.5} & 0.009 --\textgreater 0.007 \\
\hline
MoteStrain & 89 & 82.5 & 88.2 & \textbf{91} & 0.714 --\textgreater 0.592 \\
\hline
Strawberry & 95.5 & 95.6 & 87.8 & \textbf{96} & 0.063 --\textgreater 0.052 \\
\hline
CinCECGTorso & \textbf{91.5} & 64.9 & 63.2 & 87 & 0.740 --\textgreater 0.619 \\
\hline
Wine & 61.1 & 57.4 & 70.4 & \textbf{77.8} & 0.002 --\textgreater 0.002 \\
\hline
Ham & 60 & 49.5 & 71.4 & \textbf{81} & 0.440 --\textgreater 0.370 \\
\hline
SonyAIBORobotSurface1 & 64.5 & 72.5 & 71.7 & \textbf{75} & 0.423 --\textgreater 0.356 \\
\hline
Haptics & 39.5 & 38.3 & 40.9 & \textbf{55.9} & 0.430 --\textgreater 0.365 \\
\hline
ToeSegmentation2 & 80.8 & 83.8 & 80.8 & \textbf{90.8} & 0.876 --\textgreater 0.747 \\
\hline
ProximalPhalanxTW & 70.5 & 74.1 & 65.9 & \textbf{80} & 0.008 --\textgreater 0.007 \\
\hline
ChlorineConcentration & 62.5 & \textbf{64.9} & 53 & 58.5 & 0.311 --\textgreater 0.271 \\
\hline
AllGestureWiimoteY & 45.5 & \textbf{68.9} & 57.1 & 54 & 0.882 --\textgreater 0.786 \\
\hline
HouseTwenty & 68.1 & 82.3 & 83.2 & \textbf{84.5} & 0.912 --\textgreater 0.817 \\
\hline
Lightning2 & 75.4 & \textbf{80.3} & 70.5 & 78.7 & 0.554 --\textgreater 0.498 \\
\hline
AllGestureWiimoteX & 45.5 & \textbf{71.6} & 54.3 & 58.5 & 0.899 --\textgreater 0.810 \\
\hline
\tiny{GunPointMaleVersusFemale} & 99.5 & 98.4 & 93.7 & \textbf{100} & 0.052 --\textgreater 0.047 \\
\hline
ToeSegmentation1 & 68.5 & \textbf{80.3} & 64.9 & 70 & 0.929 --\textgreater 0.840 \\
\hline
Beef & 66.7 & \textbf{87.3} & 66.7 & 70 & 0.233 --\textgreater 0.214 \\
\hline
FacesUCR & 73 & \textbf{90.5} & 82.5 & 85 & 0.753 --\textgreater 0.693 \\
\hline
Rock & \textbf{64} & 48 & 44 & 60 & 0.849 --\textgreater 0.790 \\
\hline
PowerCons & \textbf{97.8} & 90 & 95 & \textbf{97.8} & 0.492 --\textgreater 0.460 \\
\hline
\tiny{DistalPhalanxOutlineCorrect} & 71.5 & \textbf{72.8} & 72.5 & 70 & 0.136 --\textgreater 0.128 \\
\hline
OSULeaf & 52 & \textbf{83.3} & 50.9 & 70 & 0.821 --\textgreater 0.772 \\
\hline
TwoPatterns & 90 & \textbf{100} & 80.2 & \textbf{100} & 0.935 --\textgreater 0.900 \\
\hline
\tiny{DistalPhalanxOutlineAgeGroup} & 62.6 & 74.8 & 69.8 & \textbf{75.5} & 0.109 --\textgreater 0.105 \\
\hline
GunPointOldVersusYoung & \textbf{100} & \textbf{100} & 91.4 & \textbf{100} & 0.048 --\textgreater 0.047 \\
\hline
BeetleFly & 75 & 63.3 & 70 & \textbf{95} & 0.946 --\textgreater 0.914 \\
\hline
ScreenType & 35.5 & 39.5 & 44 & \textbf{52.4} & 0.941 --\textgreater 0.923 \\
\hline
LargeKitchenAppliances & 50.5 & \textbf{78.4} & 49.1 & 75.5 & 0.984 --\textgreater 0.967 \\
\hline
DodgerLoopWeekend & \textbf{98.4} & 95.6 & 97.7 & \textbf{98.4} & 0.133 --\textgreater 0.132 \\
\hline
ShapeletSim & 53.9 & 62.8 & 53.9 & \textbf{65} & 0.998 --\textgreater 0.990 \\
\hline
SmallKitchenAppliances & 32.5 & 62.9 & \textbf{66.7} & 62.2 & 0.996 --\textgreater 0.989 \\
\hline
Earthquakes & 71.2 & \textbf{80} & 74 & 74.8 & 0.993 --\textgreater 0.990 \\
\hline
RefrigerationDevices & 38 & 46.1 & 40.8 & \textbf{47} & 0.993 --\textgreater 0.990 \\
\hline
Fungi & 83.9 & 79.6 & \textbf{87.1} & 85.5 & 0.000 --\textgreater 0.000 \\
\hline
FaceFour & 78.4 & \textbf{86.4} & 81.8 & \textbf{86.4} & 0.694 --\textgreater 0.706 \\
\hline
Mallat & 88 & 93.7 & \textbf{94.8} & 92 & 0.041 --\textgreater 0.042 \\
\hline
DodgerLoopGame & 87.6 & \textbf{89.1} & 78.3 & 87.6 & 0.156 --\textgreater 0.176 \\
\hline
InsectEPGRegularTrain & \textbf{100} & \textbf{100} & \textbf{100} & \textbf{100} & 0.025 --\textgreater 0.029 \\
\hline
DodgerLoopDay & \textbf{55.8} & 46.2 & 49.4 & 52.6 & 0.129 --\textgreater 0.151 \\
\hline
InsectEPGSmallTrain & \textbf{100} & \textbf{100} & \textbf{100} & \textbf{100} & 0.019 --\textgreater 0.023 \\
\hline
Coffee & \textbf{100} & \textbf{100} & 96.4 & \textbf{100} & 0.013 --\textgreater 0.016 \\
\hline
MelbournePedestrian & \textbf{93.5} & 87.7 & 71.4 & 80 & 0.122 --\textgreater 0.322 \\
\hline
OliveOil & \textbf{86.7} & 58.7 & \textbf{86.7} & 66.7 & 0.000 --\textgreater 0.002 \\
\hline
\hline
\textbf{Average} & 73.6 & 76.6 & 72.2 & \textbf{79.7} & \textbf{0.425 --\textgreater 0.330} \\
\hline
\textbf{MPCE} & 0.0832 & 0.0760 & 0.0881 & \textbf{0.0627} & ------ \\
\hline
\end{tabular}
\end{minipage}
\end{table}

\begin{figure}
\begin{center}
\includegraphics[width=0.8\linewidth]{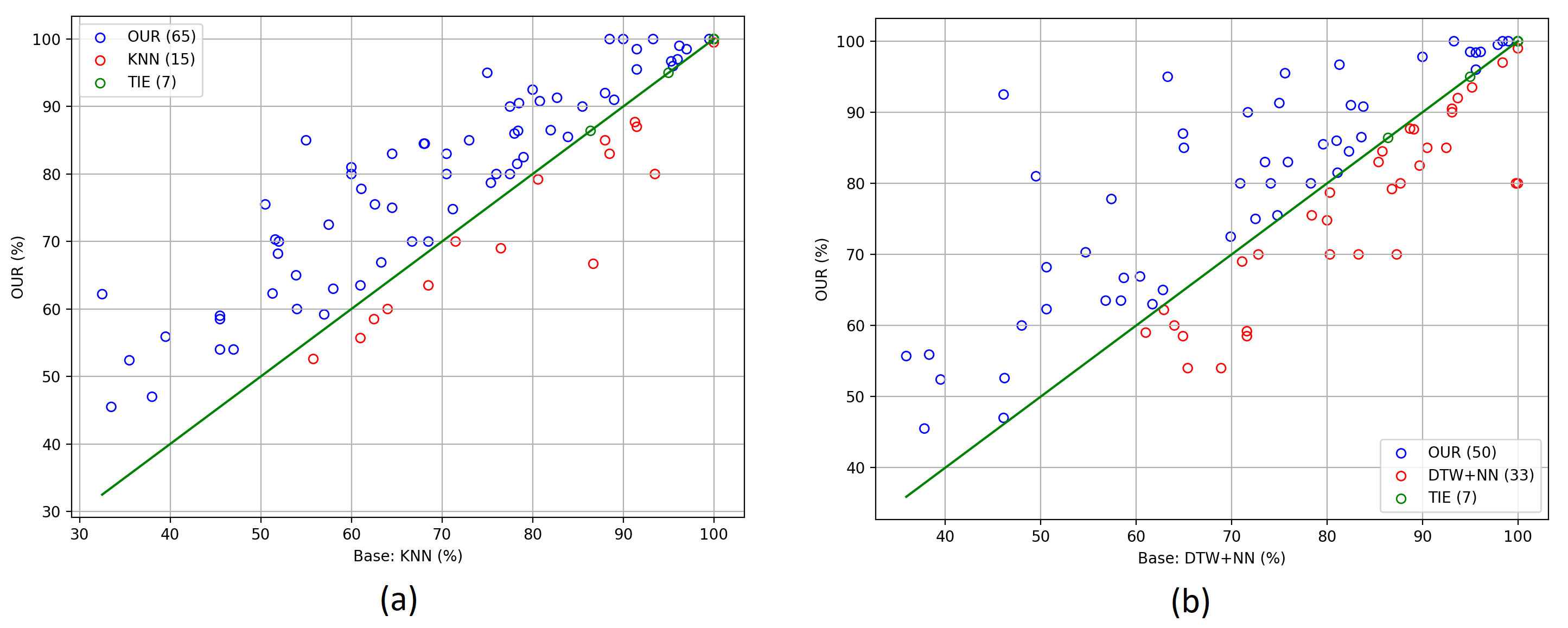}
\caption{\label{fig3-3} Scatter plot comparing our method with (a) NN and (b) DTW+NN. Each point represents a dataset, with points above the $y=x$ line indicating a win for our model (blue points) and those below showing a loss (red points).}
\end{center}
\end{figure}

Finally, Fig. \ref{fig3-3} illustrates the wins and losses of our model compared to both the NN and DTW+NN baselines. In each plot, blue points represent wins, while red points indicate losses. As shown in the figure, our model outperforms NN in 65 out of 90 tested datasets, with 15 losses. When compared to DTW+NN, our model achieves 50 wins and 33 losses. The results confirm the effectiveness of our approach against both baselines.

\subsection{Deep Network Classification}
After evaluating our method's effectiveness in enhancing the accuracy of a simple nearest neighbor classifier, this section examines its performance with a more advanced and complex classifier.

In \cite{tsc} deep learning methods for time series classification are explored, identifying \textbf{ResNet} \cite{resnet} as the best-performing model among nine top-rated approaches for UCR Archive datasets. Our method is not an alternative to ResNet but can serve as a pre-stage warper to improve the accuracy. To demonstrate this, we randomly selected 30 datasets from the previous 90 (due to computational constraints) and trained the ResNet classifier for 1500 epochs, as recommended in \cite{resnet}. Each dataset was tested twice: once in its original form and once after warping, where each test signal was warped using the model that produced the least error.

Table \ref{tab2} presents the results, showing percentage improvements in test loss average and variance for the selected datasets. These values are computed from epoch 300 to 1500 to exclude high initial variations. The results indicate a \textbf{33\%} improvement in average loss and a \textbf{54\%} reduction in variance when incorporating our warper stage. Additionally, Table 3 reports final test accuracies, revealing a \textbf{2.5\%} average accuracy improvement and a \textbf{22.7\%} reduction in MPCE (from 0.0374 to 0.0289). Notably, our approach is significantly faster than ResNet, ensuring that its integration does not introduce noticeable computational overhead.

\begin{table}
\begin{center}
\begin{minipage}{\linewidth}
\fontsize{7pt}{7pt}\selectfont
\caption{RESNET Test Loss Average and Variance percentage improvements over epochs (after epoch 300) and Accuracy comparison for 30 datasets when a warping stage with our approach is added.} \label{tab2}
\begin{tabular}{||c| c c| c c||}
\hline
\textbf{Dataset name} & \textbf{\makecell{\% Loss Avg. \\ Improvement}} & \textbf{\makecell{\% Loss Var. \\ Improvement}} & \textbf{\makecell{\% Acc. Without \\ Pre-warping}} & \textbf{\makecell{\% Acc. With \\ Pre-warping}}\\ [0.5ex] 
\hline
\hline
Birdchicken & 50.4 & 92.8 & 85 & \textbf{95} \\
\hline
BME & 81.5 & 62 & 98.7 & \textbf{100} \\
\hline
CBF & 62.8 & 21.5 & \textbf{99.4} & \textbf{99.4} \\
\hline
Coffee & 40.4 & 85.7 & \textbf{100} & \textbf{100} \\
\hline
DistalPhalanxTW & -23.6 & 35.6 & 68.3 & \textbf{71.2} \\
\hline
DodgerLoopGame & 37.4 & 97.5 & 48.8 & \textbf{51.2} \\
\hline
Earthquakes & 3.7 & -24.1 & 69.1 & \textbf{75.5} \\
\hline
ECG5000 & 8.9 & -53.4 & 93.3 & \textbf{93.6} \\
\hline
FaceFour & 14.2 & 86.1 & \textbf{95.4} & 94.3 \\
\hline
FreezerRegularTrain & 72.9 & 100 & \textbf{99.8} & 98.7 \\
\hline
GunPoint & 91.9 & 100 & 98.7 & \textbf{99.3} \\
\hline
GunPointOldVersusYoung & 99 & 100 & 97.8 & \textbf{100} \\
\hline
Herring & 34.1 & 79.9 & 60.9 & \textbf{65.6} \\
\hline
LargeKitchenAppliances & -40.8 & 21.7 & 81.1 & \textbf{90.4} \\
\hline
Lightning2 & 20.3 & 97.2 & 77 & \textbf{83.6} \\
\hline
Mallat & 5.2 & -20.6 & 91.2 & \textbf{97.4} \\
\hline
MoteStrain & 40.2 & 70.4 & 91.4 & \textbf{93.7} \\
\hline
PowerCons & 42.1 & 74.3 & 86.1 & \textbf{90} \\
\hline
ProximalPhalanxOutlineAgeGroup & -3.1 & 57.3 & 82.9 & \textbf{88.3} \\
\hline
ProximalPhalanxOutlineCorrect & 17.7 & -7.8 & 91.4 & \textbf{93.1} \\
\hline
RefrigerationDevices & -8.8 & 52 & 51.7 & \textbf{53.1} \\
\hline
SonyAIBORobotSurface1 & -30.1 & 23.2 & 93.3 & \textbf{94} \\
\hline
Symbols & 13 & -102.4 & 91 & \textbf{95.5} \\
\hline
SyntheticControl & 69.3 & 100 & \textbf{99.3} & 98.7 \\
\hline
ToeSegmentation1 & 42 & 95.9 & 96.9 & \textbf{98.7} \\
\hline
Trace & 99.7 & 100 & \textbf{100} & \textbf{100} \\
\hline
TwoLeadECG & -3.9 & 12.4 & \textbf{100} & \textbf{100} \\
\hline
TwoPatterns & 88.3 & 100 & 95.9 & \textbf{99.7} \\
\hline
UMD & 10 & 72.3 & 98.6 & \textbf{99.3} \\
\hline
Wafer & 58.5 & 99.7 & \textbf{99.8} & 99.7 \\
\hline
\hline
\hline
\textbf{MPCE} & --- & --- & 0.0374 & \textbf{0.0289} \\
\hline
\end{tabular}
\end{minipage}
\end{center}
\end{table}

\section{Conclusion}

We present a novel deep learning-based framework for MTSA, addressing a largely overlooked problem in the literature. Unlike traditional MSA methods that rely on pairwise alignments, leading to high computational complexity, our approach introduces a grouped multiple alignment algorithm that aligns all signals together. Additionally, we decompose complex non-linear warpings into simpler linear sections, ensuring a general time warping that adheres to three essential constraints. By optimizing cost functions and training procedures, our method achieves promising results in both time series classification and warped averaging.


\bibliography{sn-bibliography}

\end{document}